\documentclass[11pt,a4paper]{article}
\usepackage{times,latexsym}
\usepackage{url}
\usepackage[T1]{fontenc}

\usepackage[acceptedWithA]{tacl2018v2}

\usepackage{amssymb}

\usepackage{booktabs}
\usepackage{balance}
\usepackage{array}
\usepackage{listings}
\usepackage{graphicx}
\usepackage{subcaption}
\usepackage{multirow}
\usepackage{color, colortbl}
\definecolor{Gray}{gray}{0.9}
\usepackage{enumitem}
\usepackage{amsmath}
\usepackage{mathtools}
\usepackage{commath}
\usepackage{babel}

\usepackage[TS1,T1]{fontenc}
\usepackage{lmodern}

\usepackage{xspace,mfirstuc,tabulary}
\newcommand{\dateOfLastUpdate}{Sept. 20, 2018}
\newcommand{\styleFileVersion}{tacl2018v2}

\newif\iftaclinstructions
\taclinstructionsfalse % AUTHORS: do NOT set this to true
\iftaclinstructions
\renewcommand{\confidential}{}
\renewcommand{\anonsubtext}{(No author info supplied here, for consistency with
TACL-submission anonymization requirements)}
\newcommand{\instr}
\fi

\iftaclpubformat % this "if" is set by the choice of options

\newcommand{\TaclPapers}{Final Versions\xspace}
\else

\newcommand{\TaclPapers}{Submissions\xspace}
\fi

\title{Formatting Instructions for TACL \TaclPapers \\
(Base files: \styleFileVersion-template.tex \& \styleFileVersion.sty, dated \dateOfLastUpdate)}

\author{Lijun Lyu$^1$, Maria Koutraki$^1$, Martin Krickl$^2$, Besnik Fetahu$^{1,3}$ \\
 $^1$L3S Research Center, Leibniz University of Hannover / Hannover, Germany\\
 $^2$Austrian National Library / Vienna, Austria\\
 $^3$Amazon / Seattle, WA, USA\\
  {\sf lyu@L3S.de, koutraki@L3S.de, martin.krickl@onb.ac.at, besnikf@amazon.com} \\
}

\date{}

\newcommand*\longs{{\fontencoding{TS1}\selectfont s}}

\title{Neural OCR Post-Hoc Correction of Historical Corpora}

\begin{document}
\maketitle

\begin{abstract}
Optical character recognition (OCR) is crucial for a deeper access to historical collections. OCR needs to account for \emph{orthographic} variations, \emph{typefaces}, or \emph{language evolution} (i.e., new \emph{letters}, \emph{word spellings}), as the main source of \emph{character}, \emph{word}, or \emph{word segmentation} transcription errors. For digital corpora of historical prints, the errors are further exacerbated due to low scan quality and lack of language standardization.

For the task of OCR post-hoc correction, we propose a neural approach based on a combination of recurrent (RNN) and deep convolutional network (ConvNet) to correct OCR transcription errors. At character level we flexibly capture  errors, and decode the corrected output based on a novel attention mechanism. Accounting for the input and output similarity, we propose a new loss function that rewards the model's correcting behavior.

Evaluation on a historical book corpus in German language shows that our models are robust in capturing diverse OCR transcription errors and reduce the word error rate of 32.3\%  by more than 89\%.
\end{abstract}

\section{Introduction}
OCR is at the forefront of digitization projects for cultural heritage preservation. The main task is to \emph{identify} characters from their \emph{visual} form into their \emph{textual} representation. 

\begin{figure}[t!]
    \centering
    \includegraphics[width=0.9\columnwidth]{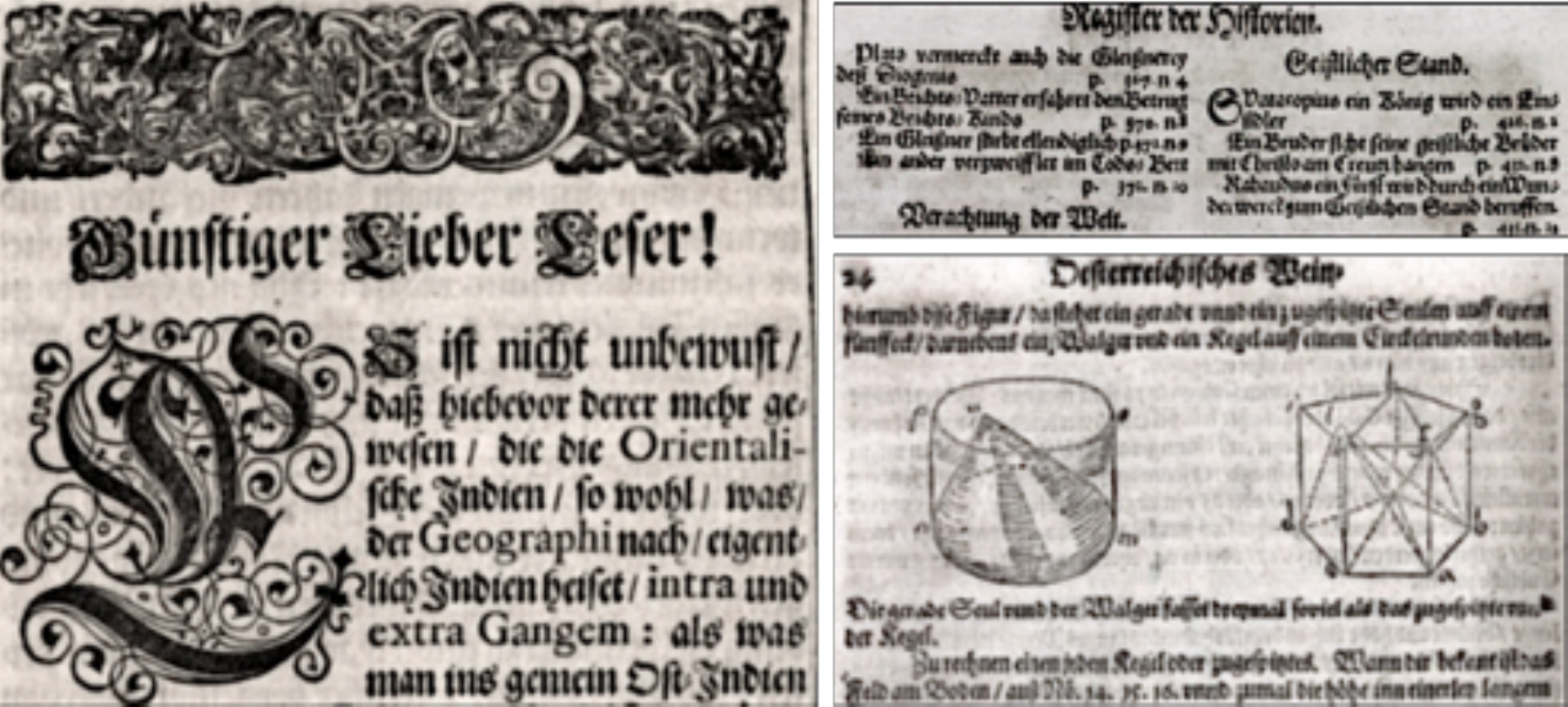}\vspace{-10pt}
	\caption{\small{Pages with coexisting typefaces (\emph{Fraktur} and \emph{Antiqua}), double columns, and images surrounded by texts.}}
    \label{fig:pageExamples}
\end{figure}

\emph{Scan quality}, \emph{book layout}, \emph{visual} character similarity are some of the factors that impact the output quality of OCR systems. This problem is severe for historical corpora, which is the case in this work. We deal with \emph{historical} books in German language from the 16th--18th century, where characters are added or removed (e.g. \emph{long s} -- \longs), word spellings change (e.g. \emph{``vnd''} vs. \emph{``und''}) that often lead to word and character transcription errors. Figure~\ref{fig:pageExamples} shows examples pages conveying the complexity of this task.

There are several strategies to correct OCR transcription errors. \emph{Post-hoc correction} is the most common  setup~\cite{dong2018multi,xu2017retrieving}. The input is an OCR transcribed text, and the output is its corrected version according to the error-free ground-truth transcription. For instance, Dong and Smith~\shortcite{dong2018multi} use a \emph{multi-input attention} to leverage \emph{redundancy} among textual snippets for correction. Alternatively, domain specific OCR engines can be trained~\cite{DBLP:conf/das/ReulSWP18}, by using manually aligned \emph{line image segments} and \emph{line text}~\cite{DBLP:journals/corr/abs-1810-03436}. However, manually acquiring such ground-truth is highly expensive, and furthermore, typically, historical corpora do not contain redundant information. Moreover, each book has its own characteristics, e.g. \emph{typeface} styles, \emph{regional} and \emph{publisher's} use of language etc.

In this work, we propose a post-hoc approach to correct OCR transcription errors, and apply it to a historical collection of books in German language. As input we have only the OCR transcription of book from their scans, for which we output the \emph{corrected} text, that we assess w.r.t the ground-truth transcription carried out by human annotators without any spelling change, language normalization, or any other form of interpretation. By considering only the textual modality for our approach, we provide greater flexibility of applying our approach to historical collections where the image scans are not available. However, note that since orthography was not standardized, there can be \emph{parallel} spellings of the \emph{`same'} word (e.g. \emph{`und'} vs. \emph{`vnd'}) within the same book, which may pose challenges for approaches that use the text modality only.% Yet, in our book corpus, parallel spellings are infrequent.

Our approach, \textbf{CR}, consists of an encoder-decoder architecture. It encodes the erroneous input text at \emph{character level}, and outputs the corrected text  during the decoding phase. Representation at character level is necessary given that OCR transcription errors at the most basic level are at character level. The input is encoded through a combination of RNN and deep ConvNet~\cite{lecun1995convolutional} networks. Our encoder architecture allows to flexibly encode the erroneous input for post-hoc correction. RNNs capture the \emph{global input context}, whereas ConvNets construct local \emph{sub-word} and \emph{word compound} structures. During decoding the errors are corrected through an RNN decoder, which at each step through an attention mechanism combines the RNN and ConvNet representations and outputs the corrected text.

Finally, since the input and output snippets are highly similar, loss functions like \emph{cross-entropy} lean heavily towards rewarding \emph{copying} behavior. We propose a custom loss function that rewards the model's ability to \emph{correct} transcription errors.

In this work, we make the following contributions:
\begin{itemize}[leftmargin=*]
\itemsep0em
	\item a data collection approach with a parallel corpus of 800k sentences from 12 books (16th--18th century) in German language;
	\item an error analysis, emphasizing the diversity and difficulty of OCR errors;
	\item an approach that flexibly captures erroneous transcribed OCR textual snippets and robustly corrects character and word errors for historical corpora.
\end{itemize}

\section{Related Work}\label{sec:related_word}

\textbf{Redundancy based. } The works in~\cite{lund2013combining, lund2011progressive, xu2017retrieving,lund2014well} view the problem of post-hoc correction under the assumption of \emph{redundant} text snippets. That is, multiple redundant text snippets are combined and under the majority voting scheme the correction is carried out. Dong et al. \cite{dong2018multi} propose a multi-input attention model, which uses redundant textual snippets to determine the correct transcription during the training phase. While there is redundancy for contemporary texts, this cannot be assumed in our case, where only the OCR transcriptions are available. Our approach can be seen as complementary to data augmentation techniques that exploit redundancy.

\textbf{Rule based Correction.} Rule based approaches compute the edit cost between two text snippets based on weighted finite state machines (WFSM) \cite{brill2000improved,dreyer2008latent,wang2014probabilistic,silfverberg2016data,farra2014generalized}. WFSM require predefined rules (\emph{insertion, deletion, etc.} of characters) and a lexicon, which is used to assess the transformations. The rewrite rules require the mapping to be done at the word and character level~\cite{wang2014probabilistic, silfverberg2016data}. This process is expensive and prohibits learning rules at scale. Furthermore, lexicons are severely affected by out-of-vocabulary (OOV) problems, especially for historical corpora. A similar strategy is followed by Barbaresi et al.~\cite{barbaresi2016bootstrapped}, who employ a spell checker to detect OCR errors and generate correction candidates by computing the edit distance. OCR transcription errors are highly contextual and there are no one-to-one mappings of misrecognized characters that can be addressed by rules (cf. Figure~\ref{fig:error_chars}).

\textbf{Machine Translation.}  Post-hoc correction can also be viewed as a special form of machine translation~\cite{kalchbrenner2013recurrent,cho2014learning, sutskever2014sequence}. For post-hoc correction of OCR transcription errors, the only reasonable representation is based on characters. This is due to the character errors and word segmentation issues, which can only be detected when encoding the input text at character level. Results from \emph{spelling correction}~\cite{xie2016neural} and \emph{machine translation} ~\cite{ling2015character,ballesteros2015improved, chung2016character, kim2016character, csahin2018character} indicate that character based models perform the best. Methods based on statistical machine translation (SMT)~\cite{DBLP:conf/lrec/AfliQWS16} use a combined set of features at word level and language models for post-hoc correction. Schulz and Kuhn~\shortcite{schulz-kuhn-2017-multi} use a multi-modular approach combining dictionary lookup and SMT for word segmentation and error correction. However, the dataset used for training is limited to books of the same topic, and requires manual supervision in terms of feature engineering. 

\textbf{Sequence Learning.} As it is shown later, character based RNN models~\cite{xie2016neural,schnober2016still} are insufficient to capture the complexity of compound-rich languages like German. Alternatively, ConvNets have been successfully applied in sequence learning~\cite{gehring2017convolutional,gehring2016convolutional}. Although the performance of ConvNet alone is insufficient for post-hoc correction, we show that their combination yields optimal post-hoc correction performance.

\textbf{OCR engines.} Slightly related are the works~\cite{DBLP:conf/das/ReulSWP18,DBLP:journals/corr/abs-1810-03436}, which retrain OCR engines on a specific domain. The assumption is that clean \emph{line scans} with the same fontface are available. In this way, the trained OCR engines are more robust in transcribing text scans of the same fontface. Figure~\ref{fig:pageExamples} shows that this is rarely the case, and many characters induce orthographic ambiguity. Furthermore, in many cases  the OCR process is unknown, with image scans being the only material available.

\section{Data Collection \& Ground-Truth}\label{sec:data_collection}

In this section, we describe our data collection efforts and the ground-truth construction process. Currently, there is no large-scale historical corpus in German language that can be used for post-hoc correction of OCR transcribed texts. The collected corpus and constructed ground-truth of more than 854k pairs of OCR transcribed textual snippets and their corresponding manual transcriptions, together with the source code are available\footnote{\url{https://github.com/GarfieldLyu/OCR_POST_DE}}.

% There is no large-scale post-hoc OCR ground-truth for historical corpora in German language. Here, we describe the data collection and ground-truth creation process. The constructed ground-truth consists of 854k OCRed textual snippets that are mapped to their manually transcribed snippets. The constructed ground-truth will be \emph{available for download} to facilitate research in this direction. The process of ground-truth construction is described below. 

\subsection{Books Corpus}\label{subsec:books_corpus}
We first describe the process behind selecting our corpus of historical books in German language. As our input textual snippets for OCR post-hoc correction we consider the publicly available historical collection of transcribed books, which are freely accessible by the  \emph{Austrian National Library} (OeNB)~\cite{onb}. The transcription of books from their image scans is done in partnership with Google Books project, which employs Google's proprietary OCR frameworks. Given that this process is an automated process, the transcriptions are not error free. 

For the ground-truth transcriptions we turn to another publicly available collection, namely \emph{Deutsches Textarchiv} (DTA)~\cite{DT}. It contains manually transcribed books based on community efforts. The transcriptions are error free and as such are suitable to be used as our ground-truth.  We consider the overlap of books present in both DTA and OeNB, providing us with the erroneous input textual snippet from OeNB and the corresponding target error-free transcription from DTA.

%To construct the ground-truth we need to \emph{map} OCRed text snippets to their corresponding manual transcriptions. As an \emph{OCR corpus}, we use the publicly available OCRed historical collections~\cite{AustrianLib} by the \emph{Austrian National Library}\footnote{\url{https://www.onb.ac.at/}}, while, for \emph{manual transcriptions} we turn to \emph{Das Deutsche Textarchiv}\footnote{\url{http://www.deutschestextarchiv.de/}}.

\begin{table*}[h!]
\resizebox{\textwidth}{!}{%
    \centering
    \begin{tabular}{l l l l l l l l l}
        \toprule
      
  ID  & Barcode  & Year & Author & Location & Layout & pages & WER & CER \\
	\midrule
	1 & Z165780108 & 1557 & H. Staden & Marburg & single column & 177 & 66.8\% & 16.3\% \\
	\midrule
	2  & Z205600207 & 1562 & M. Walther & Wittenberg & single column & 75 & 54.1\% & 12\% \\
	\midrule
	3 & Z176886605  & 1603 & B. Valentinus & Leipzig & single column & 134 & 46.8\% & 14.3\% \\
	\midrule
    4 & Z185343407 & 1607 & W. Dilich & Kassel & single column &313 &60.8\%  & 17.1\% \\
    \midrule
    5 & Z95575503  & 1616 & J. Kepler & Linz & single column w. pg. margin & 119 & 59\% & 17.2\% \\
    \midrule
    6 & Z158515308  & 1647 & A. Olearius & Schleswig & single column  & 600 & 51.8\% & 17.7\%\\
    \midrule
    7 & Z176799204 & 1652 & S. von Birken & N\"urnberg & single column & 190 & 55.9\% & 13.8\%\\
    \midrule
    8 & Z165708902  & 1672 & J. Jacob Saar & N\"urnberg & single column w. pg. margin & 186 & 33\% & 10.4\%\\
    \midrule
    9 & Z22179990X  & 1691 & S. von Pufendorf & Leipzig & single column & 665 & 32.7\% & 7.7\%\\
    \midrule
    10 & Z172274605  & 1693 & A. von Sch\"onberg & Leipzig & single column & 341 & 67.6\% & 30.5\%\\
    \midrule
    11 & Z221142405  & 1699 & A. a Santa Clara & K\"oln & single/double column w. pg. margin & 794 & 51.4\% & 16.1\%\\
    \midrule
    12 & Z124117102  & 1708 & W. Bosman & Hamburg & single column & 601 & 37.8\% & 6.7\%\\
    \bottomrule
    \end{tabular}}
    \caption{\small{Detailed book information can be accessed from the \textit{\"ONB} portal using the barcode.}}
    \label{tab:book_info}
\end{table*}

\begin{figure}[ht!]
    \centering
    \includegraphics[width=0.9\columnwidth]{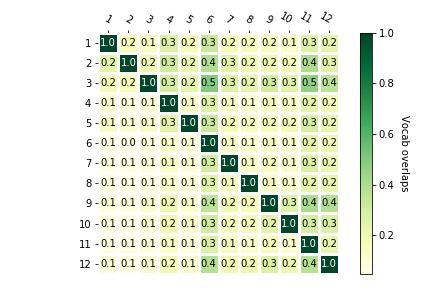}\vspace{-10pt}
    \caption{\small{Vocabulary overlaps between books.}}
    \label{fig:vocab_overlaps}
\end{figure}

Table~\ref{tab:book_info} shows our books corpus, consisting of the \emph{overlap} between these two repositories, with 12 books in \emph{German language} from the \emph{16th--18th} centuries. Understandably, considering the publication period, there is little overlap across the different books. Figure~\ref{fig:vocab_overlaps} shows the \emph{vocabulary} overlap between books, which on average is around 20\%. This presents an indicator of a corpus with high diversity and low redundancy, representing a realistic and challenging evaluation scenario for post-hoc correction.

\subsection{Ground-truth Construction}\label{subsec:ground_truth}
The constructed ground-truth consists of the mapped OCR transcribed text to their manually transcribed counterparts, resulting in a parallel corpus of OCRed \emph{input} text and the \emph{target} manually transcribed counterparts.

To construct the parallel corpus is challenging. OCR transcribed books contain all pages (e.g. \emph{content} and \emph{blank} pages), while the manually transcribed books keep only the \emph{content} pages. Furthermore, books are typically transcribed line by line by OCR systems, which often fail to detect page layout boundaries (multi-column layouts or printed margins). Therefore, accurate ground-truth construction even at page level is challenging.

\begin{figure}[ht!]
\centering
\includegraphics[width=0.85\columnwidth]{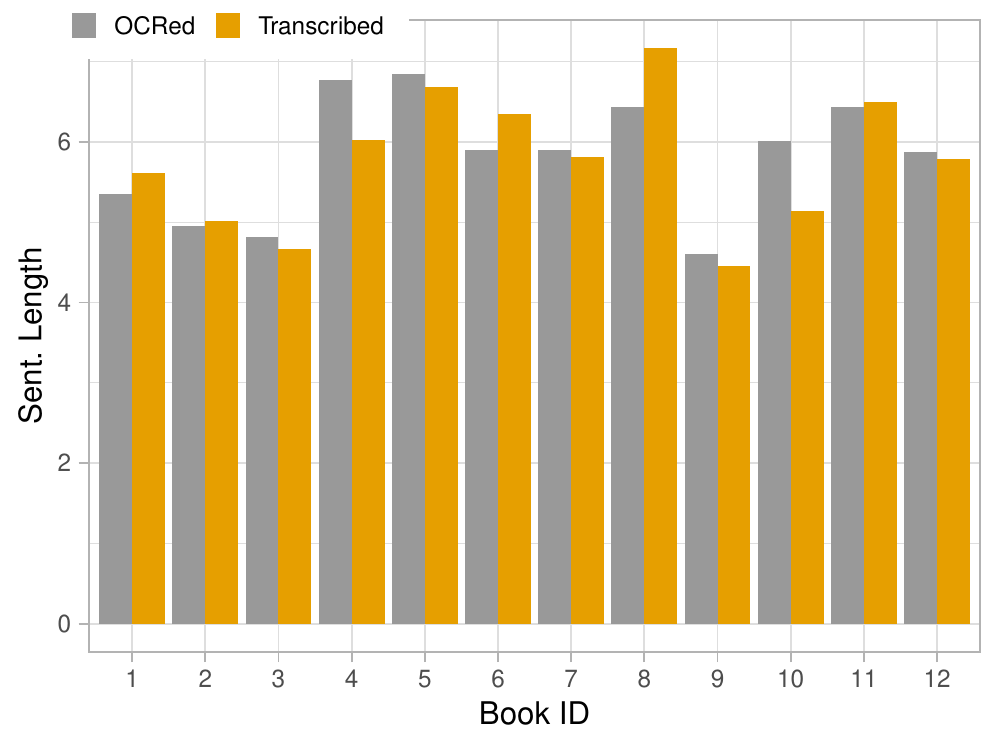}\vspace{-10pt}
\caption{\small{Sentence length distributions.}}
\label{fig:sent_dist}
\end{figure}

An important aspect is the \emph{granularity} of parallel snippets. Figure~\ref{fig:sent_dist} shows the average \emph{sentence length} distribution for OCR and manually transcribed books. We consider sentences, which are demarcated by the symbol \emph{`/'}, when this information is not available we fall back to text lines. The average sentence length is 5--6 tokens, with an average of up to 100 characters. 

Therefore, we consider snippets of \emph{5 tokens} for mapping, as longer ones (e.g paragraphs), are highly error prone. Furthermore, depending on scan quality, page content (e.g. if it contains figures or tables), the error rates from OCR transcriptions can vary greatly from page to page, making it impossible to consider lengthier snippets for the automated and large-scale ground-truth construction.

To construct an accurate ground-truth for OCR post-hoc correction, we propose the following two steps: (i) \emph{approximate matching}, and (ii) \emph{accurate refinement}. 

\subsubsection{Approximate Snippet Matching}
From the OCR transcribed books, we generate textual snippets of 5 tokens length and compute approximate matches to snippets of 5--10\footnote{Lengthier snippets are necessary due to segmentation errors, resulting in longer snippets.} tokens from the manually transcribed books. Approximate matching at this stage is required for two reasons: (i) text lines from OCR and manually transcribed books are not \emph{aligned} at line level in the books, and (ii) an exhaustive pair-wise comparison of all possible snippets of length 5 is very expensive.  

We rely on an efficient technique known as \emph{locality sensitive hashing} (LSH)~\cite{rajaraman2011mining} to put textual snippets that are \emph{loosely} similar into the same bucket, and then based on the Jaccard similarity determine the highest matching pair. The hashing signatures and the Jaccard similarity are computed on \emph{character tri-grams}.

The resulting mappings are not error free, and often contain \emph{extra} or \emph{missing} words. Such errors are introduced often due to the OCR engines breaking over the multi-column layouts of books, inclusion of table/figure captions, word segmentation errors (under or over segmentation). Snippets from OCR transcriptions that do not have a matching above a threshold ($<0.8$) are dropped. 

\textbf{Matching Coverage:} Finally, to ensure that our ground-truth construction approach does not severely affect coverage of the matched pairs, we conduct a manual analysis of two books with different layouts (books ID 6 and 11 cf. Table~\ref{tab:book_info}) for 10 randomly selected pages from each book. For book 6, which has good scan quality, for snippets of 5 tokens, we are able to find a relevant match from the manual transcription on average for 270 out of 300 snippets per page. The dropped snippets in absolute majority of the cases consist of footnotes or page headings. In the case of book 11, which has a bad scan quality and is of double column layout, from 400 snippets, only 200 have a match. Upon inspection, we find that this is mostly due to the erroneous transcription by OCR systems, which mistakenly merges lines from different columns into a single line. These snippets are corrupted, and cannot be matched to snippets extracted from the manually transcribed books.

\subsubsection{Accurate Refinement}

The main issue with the approximate matching through LSH, is that there are \emph{extra} words appearing at the head or end of either the input or output snippets. The extra tokens stem mostly from snippets that match lengthier or shorter ones due to word segmentation errors. Such \emph{additional/missing} words are not desirable, and thus, in this stage we refine the above snippet mappings.
\begin{figure}[ht!]
    \centering
    \includegraphics[width=1.0\columnwidth]{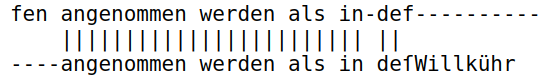}\vspace{-10pt}
    \caption{Ideal textual snippet alignment. Gap characters `-' and additional characters are removed from both sides.}
    \label{fig:alignment}
\end{figure}
We perform a \textit{local pairwise sequence alignment}~\cite{lps_package} that finds the best matching local sub-snippets. The remaining extra characters are removed, e.g tokens \emph{`fen'} and \emph{`Willk\"uhr'} are removed as they are not part of a local alignment.

\section{Data Analysis}\label{sec:data_analysis}
Based on a \emph{manual} analysis of a random sample of 100 snippet pairs taken from each book from our ground-truth, and analyze the various OCR transcription error types.

This is a crucial step towards developing post-hoc correction models in a systematic manner. OCR errors are highly contextual and are dependent on several factors, and as such there are no one-to-one rules that can be used to correct OCR errors. Furthermore, these errors are increased when dealing with historical corpora, as fontfaces, book layouts and language use are highly unstandardized. 

We differentiate between the following errors: (i) \textbf{over-segmentation} is an error when multiple words are merged into one, (ii) \textbf{under-segmentation} when a word is split into two, and (iii) \textbf{word error}, typically caused by \emph{misrecognized characters}, converting it into an \emph{invalid word} or changing its meaning to a \emph{different valid word}.

\subsection{Error Types and Distribution}
% \textbf{Overall Error Distribution.} The WER and CER rates across books are highly variable, with some books having up to 67\% of WER. These varying rates can be attributed to the different book characteristics.

\begin{figure}[t!]
    \centering
    \includegraphics[width=0.95\columnwidth]{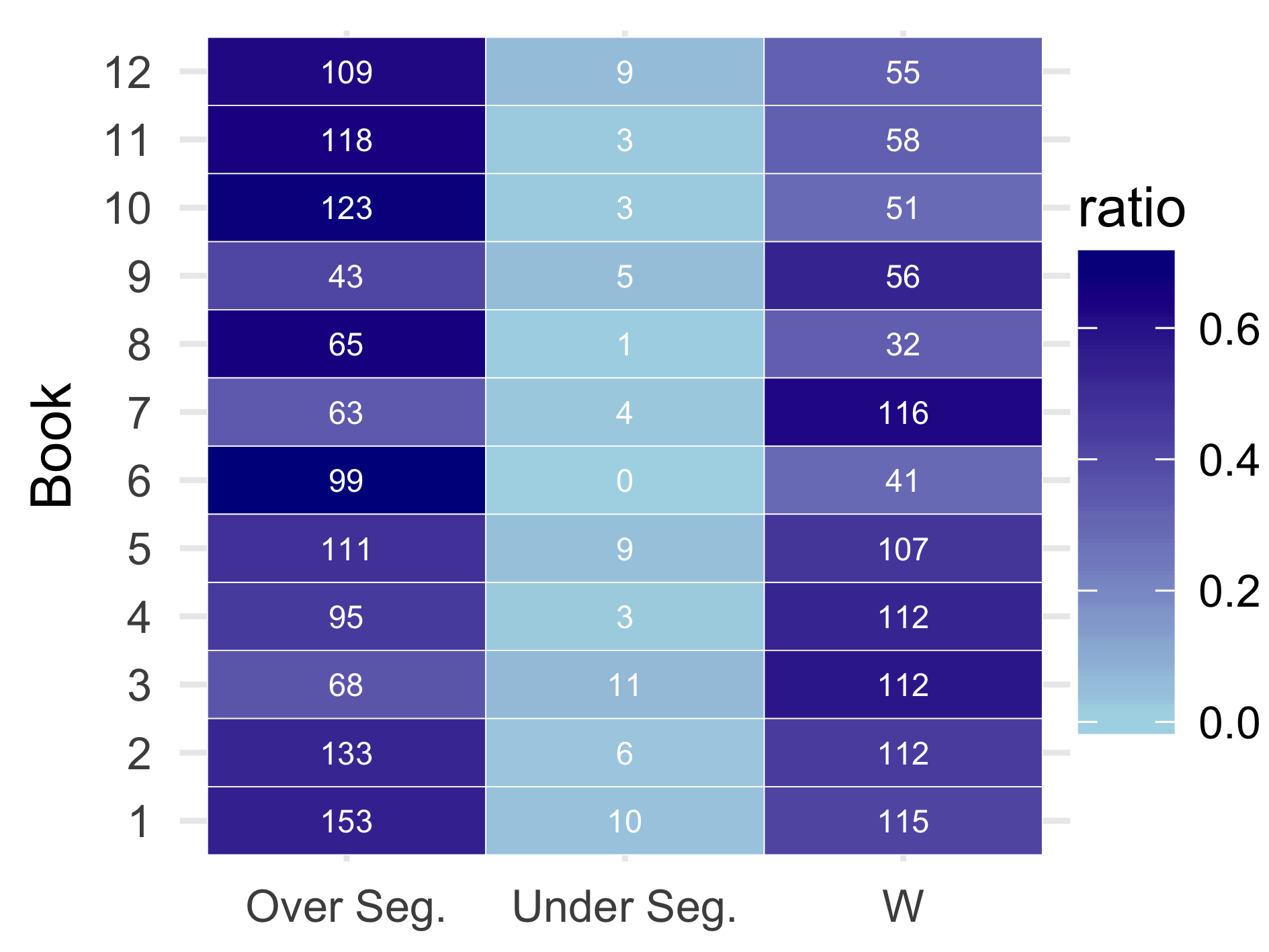}\vspace{-10pt}
    \caption{\small{Books' OCR error type distribution.}}
    \label{fig:error_type_dist}
\end{figure}

Figure~\ref{fig:error_type_dist} shows an overview of the \emph{error types} for the different books in our corpus.

\textbf{Over-segmentation} is one of the most common OCR transcription errors with 54\% of the cases. The errors often arise due to OCR systems misrecognizing spaces between words and characters in a word, as these are often not clearly distinguishable. These errors are challenging since the words may represent valid words, which is even more challenging problem for \emph{compound} rich languages like German.

\textbf{Under-segmentation} errors are less common (with 3\%), and are mostly due to line-breaks and book layouts. 

\textbf{Word-errors} represent the second most frequent OCR error category with 43\%. These errors are often caused due to the \emph{orthographic} visual similarity between characters, thus, resulting in invalid words or changing the word's meaning altogether. Other relevant factors are the scan quality or book layouts.

\begin{figure}[t!]
    \centering
    \includegraphics[width=0.95\columnwidth]{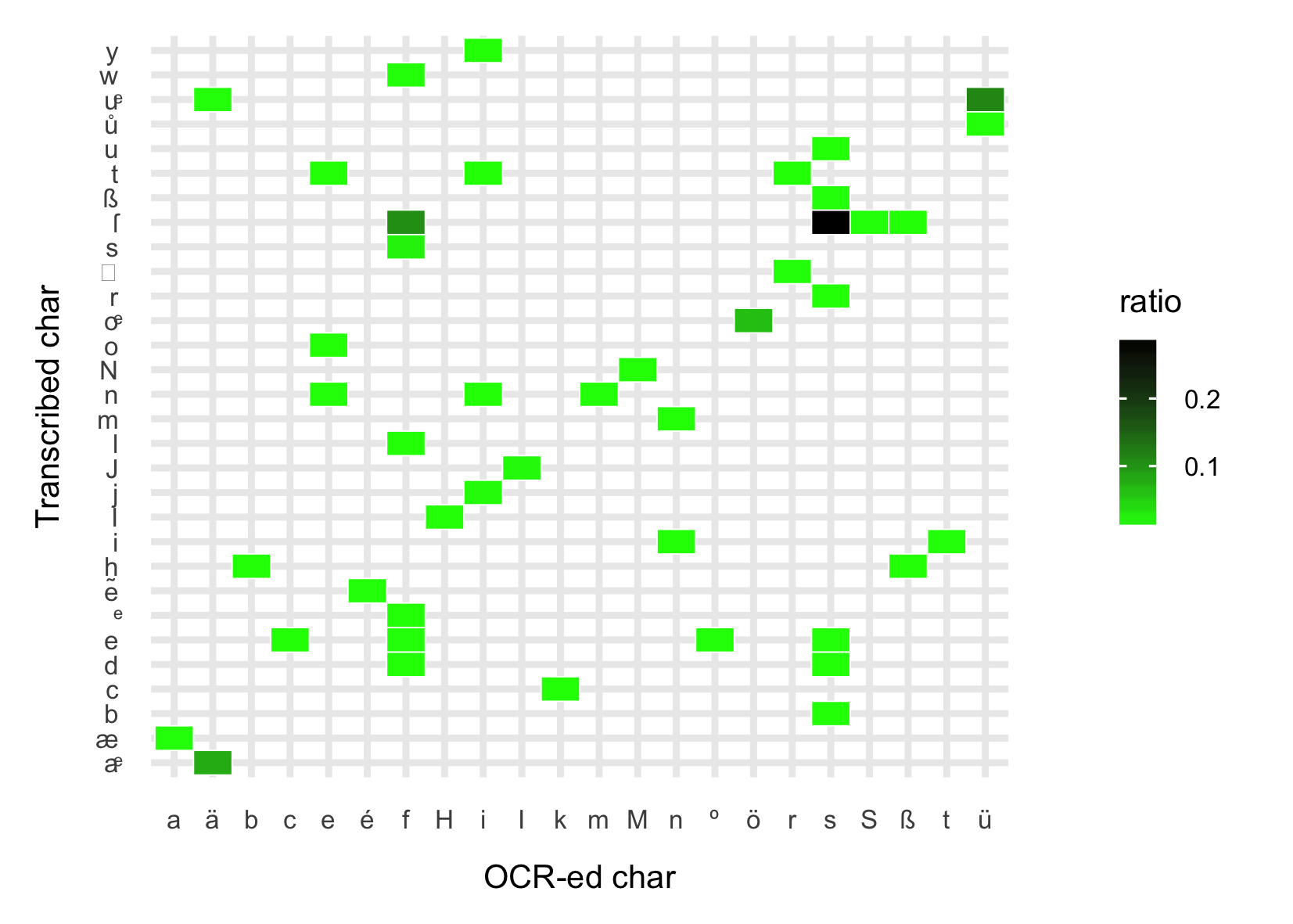}\vspace{-10pt}
    \caption{\small{The misrecognized character (x--axis) and their valid transcriptions (y--axis).}}
    \label{fig:error_chars}
\end{figure}

Figure~\ref{fig:error_chars} shows that word errors are contextual, with no simple mappings between misrecgonized characters. An indicator that they are not solely due to visual character similarity, as they are often misrecognized to completely different characters.

\section{Neural OCR Post-Hoc Correction}\label{sec:approach}

\begin{figure}[ht!]
	\centering
	\includegraphics[width=1.0\columnwidth]{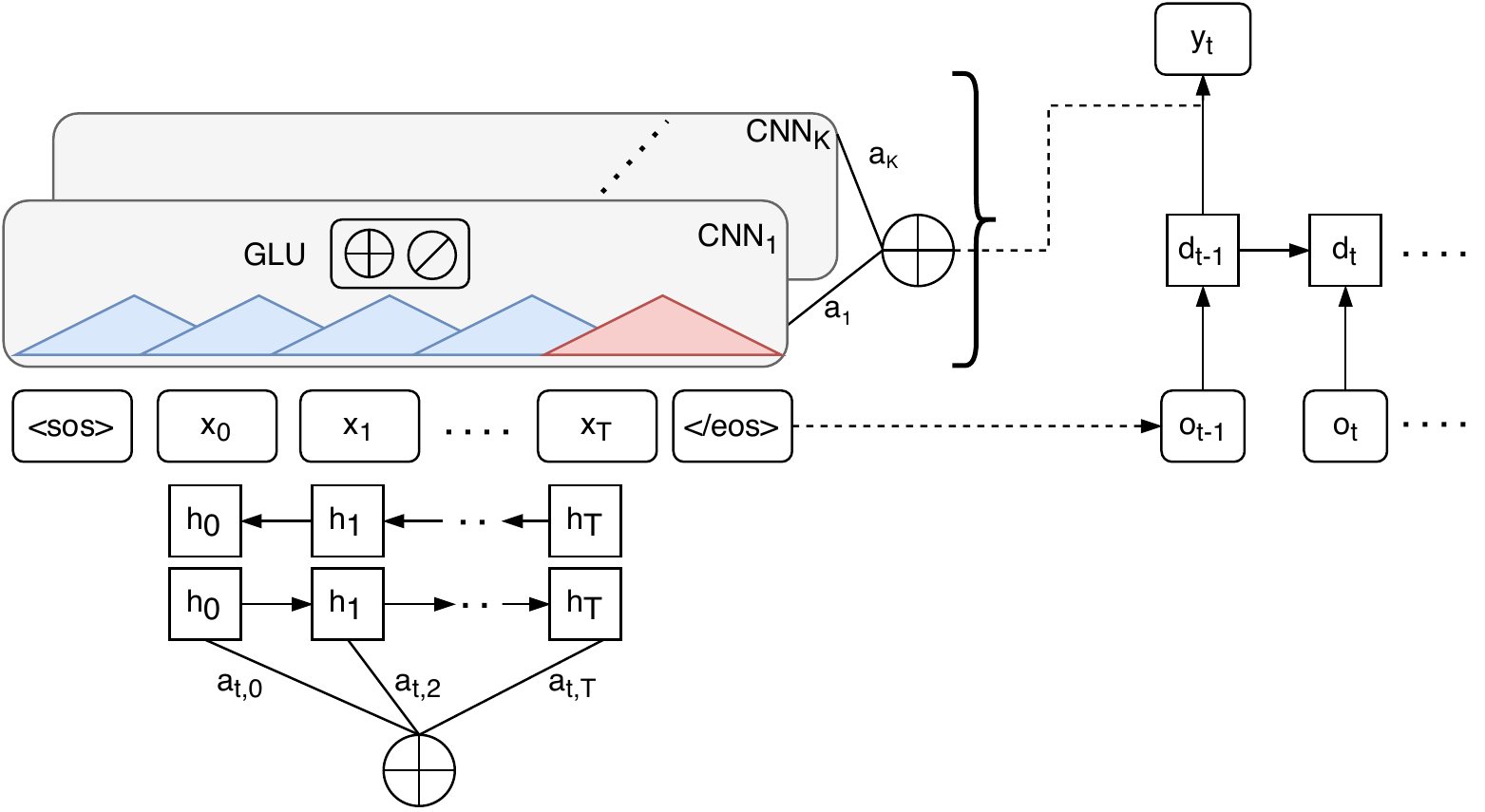}\vspace{-10pt}
	\caption{\small{Approach overview.}}
	\label{fig:approach}
\end{figure}

Figure~\ref{fig:approach} shows an overview of our encoder-decoder architecture for post-hoc OCR correction. At its core, the encoder combines RNN and deep ConvNets for representation of the \emph{erroneous} OCR transcribed snippets at \emph{character} level. During the decoder phase an RNN model corrects the errors one character at a time by employing an attention mechanism that combines the encoder representations, a process repeated until an end of a sentence is encountered.

\subsection{Encoder Network}\label{subsec:encoders}

We encode the erroneous OCR snippets at character level for three reasons. First, word representation is not feasible due to \emph{word errors}. Second, only in this way we can capture \emph{erroneous characters}. Finally, we avoid \emph{out-of-vocabulary} (OOV) issues, as there are no vocabularies for historical corpora.

The encoder consists of a RNN and a deep ConvNet network. The intuition is that, RNNs capture the \emph{global context} on how OCR errors are situated in the text, while deep ConvNets capture and enforce \emph{local sub-word/phrase} context. This is necessary for \emph{word segmentation} errors, which might bias RNN models towards more frequent tokens (e.g. \emph{``alle in''} vs. \emph{``allein''})\footnote{Both tokens are correct, with the first being more frequent}.

\subsubsection{Recurrent Encoder}\label{subsec:rnn encoder}

First, we apply a bidirectional LSTM~\cite{hochreiter1997long} that reads the erroneous OCR snippet $X=(x_1,\ldots,x_T)$, encoding it into $h_T = \left[ \overrightarrow{LSTM}\left(X\right), \overleftarrow{LSTM}\left(X\right) \right]$. 

We use recurrent models due to their ability to detect \emph{erroneous} characters and to capture the \emph{global context} of the OCR errors. In \S~\ref{sec:data_analysis} we showed that for most of the \emph{erroneous} characters, the target \emph{transcribed characters} vary, a variation that can be resolved by the general context of the snippet. 

Finally, we will use $h_T$ to conditionally initialize the decoder, since the \emph{input} and \emph{output} snippets are highly similar.

\subsubsection{Convolutional Encoder}
57\% of errors are \emph{word segmentation} errors. Often, these errors have a local behavior, such as merging or splitting words. While in theory this information can be captured by the RNN encoder, we notice that they are biased towards frequent sub-words in the corpora with tokens being wrongly split or merged. 

We apply deep ConvNets to capture the \emph{local context} (i.e. \emph{compound} information) of tokens. ConvNets through their kernels limit the influence that characters beyond a token's context may have in determining whether the subsequent decoded characters forming a token should be split or merged. 

We set the kernel size to 3 and test several configurations in terms of ConvNet layers, which we empirically assess in \S~\ref{subsec:ablation_study}. Since we are encoding the OCR input at character level, determining the right granularity of representation is not trivial. Hence, the multiple layers $l$ will flexibly learn from \emph{fine} to \emph{coarse} grained representation of the input. The learnt representation at layer $l$ is denoted as $h^l = \left(h{_1}{^l}, \ldots, h{_T}{^l}\right)$. In between each of the layers, we apply \emph{non-linearity} such as gated linear units~\cite{dauphin2017language} to control how much information should pass from the bottom to the top layers.

% The output of the $l$-th layer is denoted as $h^l = \left(h{_1}{^l}, \ldots, h{_T}{^l}\right)$. The embedding input is linearly transformed into the same dimension as the convolution layer before feeding to it. We apply non-linearity at each layer to control how much information should be kept. We apply the gated linear units (GLU)~\cite{dauphin2017language}, and residual connection at each layer.

\subsection{Decoder Network} 

The decoder is a single LSTM layer, which generates the corrected textual snippet a character at a time. We \emph{initialize} it with the last hidden state from the BiLSTM encoder $h_T$, that is $o_1=h_T$ in Equation~\ref{eq:decoder}, which biases the decoder to generate sequences that are similar to the input text.
\begin{align}\label{eq:decoder}\small
	p\big(o_i|o_{i-1},\ldots, o_1, \mathbf{x}\big) = g\big(o_{i-1}, d_{i}, c_i\big)
\end{align}
where $d_i$ is the current hidden state of the decoder, and $o_{i-1}$ represents the previously generated character. $c_i$ is the context vector from the encoded OCR input snippe, which combines the RNN and deep ConvNet input representations through a multi-layer attention mechanism which we explain below.

\subsubsection{Multi-layer Attention}\label{subsec:multi_attention}
Using jointly RNNs with deep ConvNets as encoders allows for greater flexibility in capturing the complexities of OCR errors. Furthermore, the multi-layers of the ConvNets capture from fine to coarse grained local structures of the input. To harness this encoding flexibility, we compute the context vector $c_i$ for each decoder step $d_i$ as following.

First, for each decoder state $d_i$ at step $i$, we compute the weight of the representations computed by the deep ConvNet at the different layers. The weights, computed in Equation~\ref{eq:attention_weights} correspond to the \emph{softmax} scores, which are computed based on the dot product between $d_i$ and the hidden layers $h_j^l$ from the $l$ layers of the ConvNet. 
\begin{align}\label{eq:attention_weights}\small
    a^{l}_{ij} = \frac{\exp(e^{l}_{ij})}{ \sum_{k=1}^{T}\exp(e_{ik}^{l})};\; e_{ij}^{l} = d_{i} \cdot h_{j}^{l}
\end{align}

At each layer $l$ in the ConvNet encoder, the attention weights assess the importance of the representations at the different granularity levels in correcting the OCR errors during the decoder phase. To compute $c_i$, we combine the RNN and deep ConvNet representations, as scaled by the attention weights as following:
\begin{align}\label{eq:context_vector}\small
    c_{i} = \frac{1}{L}\sum_{l=1}^{L}\sum_{j=1}^{T}a_{ij}^{l}\cdot[h_{j}^{l}, h_j]
\end{align}

% Empirically, we find that the lower ConvNet layers perform slightly better than the upper layers. Averaging through all the layers does not yield a significant improvement in terms of post-hoc OCR correction ability. One possible explanation for such a behavior can be attributed to two factors. First, the length of the input sequence is limited to 5 tokens, hence, coarse grained representations at upper layers do not necessarily provide any additional information compared to the lower levels. Second, the RNN encoder already captures the global context of the input snippet, and as such the upper ConvNet layers do not provide additional information for the decoder in post-hoc OCR correction.

\subsection{Weighted Loss Training}\label{subsec:loss_section}

Conventionally, encoder-decoder architectures are trained using the cross-entropy loss, $\mathcal{L} = -P_{tgt} \cdot \log P_{pred}$, with $P_{tgt}$ and $P_{pred}$ being the \emph{target} and the \emph{predicted} probability distributions of some discrete vocabulary. 

For OCR post-hoc correction, cross-entropy does not properly capture the nature of this task. Models are biased to simply \emph{copy} from input to output, which in this task represent the majority of cases. In this way, failure at correcting erroneous characters diminish, as all time-steps are treated equally. We propose a weighted loss function that rewards higher models for their \emph{correcting} behavior. The modified loss function is shown in Equation~\ref{eq:loss_function}.
\begin{align}\label{eq:loss_function}
    \mathcal{L}' = \mathcal{L} \cdot \big(1 - \lambda P_{src}\cdot P_{tgt}\big); \;\; 0 < \lambda < 1   
\end{align}

The new loss function combines the cross-entropy loss $\mathcal{L}$ and an additional factor that considers the source and target characters. The second part of the equation captures the amount of desirable \emph{copying} from input to output. If the input and output characters are the same, then $P_{src} \cdot P_{tgt}$ yields 1, otherwise 0, where $P_{src}$ and $P_{tgt}$ are one-hot character encodings of the input and output snippets. $\lambda$ controls by how much we want to dampen this behavior. $\mathcal{L}'$ rewards higher the model's ability to correct erroneous sequences.

\section{Experimental Setup}\label{sec:experiment}

In this section, we introduce the experimental setup and the competing methods for the task of post-hoc OCR correction. %The evaluation setup, along with the code and datasets will be publicly available for download.

\subsection{Evaluation scenarios}
According to our error analysis in \S~\ref{sec:data_analysis} and the highly diverging vocabularies across books (cf. \S~\ref{subsec:books_corpus}), we distinguish two evaluation scenarios. Here we use part of the ground-truth, where we select instances by first sampling pages from the books, namely the instance pairs coming from the sampled pages. 

We assess the performance of models for two significant factors that may impact their correction behavior: (i) \emph{eval--1} assesses the model's post-hoc correction behavior on unseen OCR transcription errors related to the book source and publication date, and diverging book content (cf. Figure~\ref{fig:vocab_overlaps}), and (ii) \emph{eval--2} tests the impact on correction performance when models have encountered all OCR errors based on random sampling.

% by training and testing on different centuries, we show the model's ability to correct possibly unseen OCR transcription errors, and (ii) \emph{eval--2}, by randomly splitting testhow do models behave when all OCR transcription errors

%, where we use two different sets from the generated ground-truth with similar topics in \S~\ref{subsec:ground_truth} for training and testing.

\paragraph{\textbf{eval--1}:} We split the data along the temporal axis, with training instances coming from books from the 16th and 18th centuries, and test instances from the 17th century. This scenario is challenging as there are diverging error types due to scan quality, and other orthographic variations related to the publishers and other book characteristics. The 17th century books have more diverse errors, as there are more books, and the initial OCR transcription error rates are higher.

We use 70\% of the data for training, and 10\% and 20\% for validation and testing, with 269k, 27k and 89k instances respectively. 

% Since the Book 12 from the 18th century contains more instances, for the test instances to account to 10\% of the overall instances in \emph{eval--1}, we downsample them to 89k instances.

\paragraph{\textbf{eval--2}:} We randomly construct the training, validation, and testing splits, thus, ensuring that the models have observed all error types, which should result in better post-hoc correction behavior. Furthermore, contrary to \emph{eval--1}, where the splits are dictated by the publication date of the books, in this case, we use slightly different splits for training, validation and testing. We use 65\%, 10\%, and 25\%, for training, validation, and testing, respectively. The absolute number is 417k, 42k and 166k respectively.

\subsection{Evaluation Metrics}\label{subsec:eval_metrics}
To assess the post-hoc correction performance of the models, we use standard evaluation metrics for this task: (i) word error rate (\textbf{WER}), and (ii) character error rate (\textbf{CER}). The error rates measure the number of word/character \emph{substitutions}, \emph{insertions}, and \emph{deletions}, normalized by the \emph{total length} of the transcribed sequence, in characters for \textbf{CER} and number of words for \textbf{WER}.

% We distinguish the following operations a model can undertake: (i) \textbf{s} and \textbf{s'} refer to the action of \emph{correctly} and \emph{incorrectly} segmenting an input token, respectively; (ii) \textbf{m} and \textbf{m'} refer to the action of \emph{correctly} and \emph{incorrectly} merging a token, and (iii) \textbf{s} and \textbf{s'} refer to the action of \emph{correctly} and \emph{incorrectly} replacing a character in a token.

\subsection{Baselines}
In the following we describe the approaches we compare against. In all cases, the input is represented at character level with 128 embedding dimensions. The cell units, i.e. LSTMs and ConvNets, are of 256 dimensions.

\paragraph{\textbf{CH}:} Xie et al.~\shortcite{xie2016neural} use an RNN model for \emph{spelling correction}, a task slightly similar to OCR post-hoc correction. Yet, the error types and their distribution are of a different nature. CH is a standard attention based encoder-decoder~\cite{bahdanau2014neural}, which corresponds to our CR model without ConvNets and the custom loss function.

\paragraph{\textbf{CH}$_\lambda$:} To assess the impact of the introduced loss function, we train CH with the custom loss (cf. \S~\ref{subsec:loss_section}). The optimal $\lambda$ is set based on the validation set. This presents the ablated model of our approach CR without the ConvNet encoder and mult-layer attention.

\paragraph{\textbf{PB:}} Cohn et al.~\shortcite{DBLP:conf/naacl/CohnHVYDH16} propose a symmetric attention mechanism for RNN based encoder-decoder models. That is, encoder and decoder timesteps are strongly aligned. A similar alignment between input and output is expected for this task.

\paragraph{\textbf{Transformer:}} By pretraining on large corpora, Transformers have~\cite{vaswani2017attention} achieved the state-of-the-art results in various NLP tasks. In our case, pretraining on historical corpora is not possible due to the scarcity of such data, while pretraining on contemporary German corpora did not show any improvement. The self-attention mechanism is highly flexible in capturing intra-input and input-output dependencies, which is very important for post-hoc correction. We use the implementation in~\cite{transformer_keras} with 3 layers and 8 attention heads, and 512 dimensions for the output model, and encode input at character level.

\paragraph{\textbf{Other approaches:}} ConvSeq~\cite{gehring2017convolutional}, part of our encoder network, yields performance below all the other competitors, hence, we do not include its results here. Similarly, rule-based models based on FST~\cite{silfverberg2016data} yield poor performance. We believe this is due to the inability to establish one-to-one mapping of rules for correction, and the requirement for valid word vocabularies.

\subsection{CR: Approach Configuration}
For our approach \textbf{CR}, based on a validation set, the number of ConvNet layers is set to $k=1$ and $k=3$, and set $\lambda=0.3$ and $\lambda=0.1$, for \emph{eval--1} and \emph{eval--2}, respectively.

\section{Evaluation}\label{sec:evaluation}
In this section, we provide a detailed evaluation discussion and discuss limitations. 
\begin{enumerate}[leftmargin=*]
\itemsep0em
    \item Post-hoc correction evaluation as measured through WER and CER metrics.
    \item Ablation study for our approach CR.
    \item Performance of CR for post-hoc correction at page level.
    \item Robustness and generalizability of our approach for post-hoc correction.
    \item CR model behavior error analysis. 
\end{enumerate} 

% Secondly, we present an ablation study of our approach, followed with a deep analysis of our model's performance at page level, measured in terms of the accuracy of the actions that are undertaken to transform the erroneous input to its target form. Finally, we present an error analysis 

\subsection{Post-Hoc OCR Error Correction}\label{subsec:model_performance}

All post-hoc OCR correction approaches under comparison reduce significantly the amount of OCR errors. Tables~\ref{tab:correction_result_eval_1} and \ref{tab:correction_result_eval_2} provide an overview of the performance as measured through WER and CER metrics. 

\paragraph{\textbf{eval--1.}} Table~\ref{tab:correction_result_eval_1} shows the results for competing approaches for the \emph{eval--1} scenario. This scenario mainly shows how well the models generalize in terms of language evolution, where instances come from books written in a different century. Noted that, apart from the \emph{temporal dimension}, another important aspect is that of \emph{publisher's} specific attributes. Dependent on the publisher, there are orthographic variations, vocabulary, and other stylistic features, such as font-face etc. 

\begin{table}[th!]
    \centering\small
    \resizebox{0.9\columnwidth}{!}{%
    \begin{tabular}{l l l }
        \toprule
 & \textbf{WER} & \textbf{CER} \\
	\midrule
	OCR & 33.3 & 6.1\\
	CH & 7.64 ($\blacktriangledown 77\%$) &  2.79 ($\blacktriangledown 54\%$) \\
	CH$_{\lambda=0.4}$ & 7.46 ($\blacktriangledown 78\%$) &  2.53 ($\blacktriangledown 59\%$)\\
	PB & 11.45 ($\blacktriangledown 66\%$) &  3.05 ($\blacktriangledown 50\%$)\\
	Transformer & 8.11 ($\blacktriangledown 76\%$) & 2.24 ($\blacktriangledown 63\% $)\\
	CR & \textbf{5.98} ($\blacktriangledown 82\%$)$^{*}$ & \textbf{2.07} ($\blacktriangledown 66\%$)$^{*}$\\
    \bottomrule
    \end{tabular}}
    \caption{\small{Correction results for \emph{eval--1}. CR achieves \emph{highly significant} ($^*$) improvements over the best baseline CH$_\lambda$.}}
    \label{tab:correction_result_eval_1}
\end{table}

In principle, low WER translates into fewer word segmentation (WS) errors, with WS errors being some of the most frequent errors (cf. Figure~\ref{fig:error_type_dist}). Hence, reducing WER is critical for post-hoc OCR correction models. Our model, \textbf{CR}, achieves the best performance with the lowest score of WER=5.98\%. This presents a relative decrease of $\Delta=82\%$ compared to the WER in the original OCR text snippets. In terms of CER we have a relative decrease of $\Delta=66\%$, namely with CER=2.07\%. 

Comparing our approach CR against CH$_\lambda$ (the best competing approach in \emph{eval--1}), we achieve highly significant ($p < .001$) lower WER and CER scores, as measured according to the non-parametric \emph{Wilcoxon signed-rank} test with correction\footnote{We test for \emph{normality} of distributions, and conclude that the produced WER and CER measures do not follow a normal distribution}. For WER and CER, CR compared to CH$_\lambda$ obtains a relative error reduction of 21.7\% and 25.8\%, respectively. This shows, that ConvNets allow for flexibility in capturing the different constituents of a word compound, that in turn may result in either \emph{over} or \emph{under} segmentation error.

Against the other competitors the reduction rates are even greater. Transformers with the lowest CER among the competitors, yet, compared to CR its CER has a 8\% relative increase. PB, performs the worst, mainly due to the character shifts (left or right) incurred due to word segmentation errors. Thus, strictly enforcing the attention mechanism along \emph{very close} or the \emph{same} positions in the encoder-decoder results in sub-optimal post-hoc OCR correction behavior.

\paragraph{\textbf{eval--2.}} Table~\ref{tab:correction_result_eval_2} shows the results for the \emph{eval--2} scenario. Due to the randomized instances for training and testing, the models have greater ability in correcting OCR errors. Contrary to \emph{eval--1} where the models were tested on instances coming from later centuries, in this scenario, the models do not suffer from \emph{language evolution} aspects and other \emph{book} specific characteristics. Therefore, this presents an easier evaluation scenario.

\begin{table}[th!]
    \centering\small
    \resizebox{0.9\columnwidth}{!}{%
    \begin{tabular}{l l l}
        \toprule
 & \textbf{WER} & \textbf{CER}\\
	\midrule
	OCR & 32.3 & 5.4 \\
	CH & 4.08 ($\blacktriangledown 87\%$) & 1.32 ($\blacktriangledown 76\%$)\\
	CH$_{\lambda=0.3}$ & 4.09 ($\blacktriangledown 87\%$)& 1.35 ($\blacktriangledown 75\%$)\\
	PB & 9.21  ($\blacktriangledown 71\%$)& 1.93 ($\blacktriangledown 64\%$)\\
	Transformer & 4.50 ($\blacktriangledown 86\%$) & \textbf{1.07} ($\blacktriangledown 80\%$)$^*$\\
	CR & \textbf{3.59} ($\blacktriangledown 89\%$)$^*$ & 1.31 ($\blacktriangledown 76\%$)\\
    \bottomrule
    \end{tabular}}
    \caption{\small{Correction results for \emph{eval--2}. CR obtains \emph{highly significant} ($^*$) improvements over the best baseline CH$_\lambda$ for WER, while Transformer has significantly the lowest CER.}}
    \label{tab:correction_result_eval_2}
\end{table}
Here too the models show a similar behavior as for \emph{eval--1}. The only difference in this case being that our approach CR does not achieve the best CER reduction rates. While, CR obtains highly significant lower ($p<.001$) WER rates than the Transformer. On the other hand, Transformer achieves the best CER rates among all competitors ($p<.001$). The significance tests are measured using the non-parametric Wilcoxon signed-rank test. 

This presents an interesting observation, showing that Transformers are capable in learning all the complex cases of character errors. This behavior can be attributed to their capability in learning complex intra-input and input-output dependencies. However, in terms of WER, we see that a large reduction is achieved through ConvNets in CR, yielding the lowest WER rates, with a relative decrease of 89\% in terms of WER. This conclusion can be achieved when we inspect CH$_\lambda$, which is the ablated CR model without ConvNet encoders.

\subsection{Ablation Study}\label{subsec:ablation_study}
In the ablation study we analyze the impact of the varying components introduced in \textbf{CR}.

\textbf{ConvNet Layers.} The number of layers provides different levels of abstractions in encoding the OCR input. Table~\ref{tab:ablation_layers} shows CR's performance with varying number of layers trained using the \emph{standard} cross-entropy loss. Increasing the number of layers for $k>5$ does not yield performance improvements. We note that for the different evaluation scenarios, the number of necessary layers varies. For instance, in \emph{eval--2} the number of optimal layers is 3. This can be attributed to the higher diversity of errors in the randomized validation instances, and thus, the need for more layers to capture the OCR errors.
\begin{table}[h!]
    \centering\small
    \resizebox{0.9\columnwidth}{!}{%
    \begin{tabular}{l l l | l l }
        \toprule
        & \multicolumn{2}{c}{\emph{eval--1}} & \multicolumn{2}{c}{\emph{eval--2}}\\
        \midrule
 & \textbf{WER} & \textbf{CER} & \textbf{WER} & \textbf{CER}\\
	\midrule
	CR$_{k=1}$ &\textbf{6.18} & \textbf{2.15} & 3.72 & 1.29\\
	CR$_{k=2}$ & 6.46 &  2.30 & 4.18 & 1.46\\
	CR$_{k=3}$ & 6.47 & 2.26 & 3.61 & \textbf{1.26}\\
	CR$_{k=4}$ & 6.93  &  2.51  & \textbf{3.54}  & 1.31 \\
	CR$_{k=5}$ & 6.63 & 2.40 & 3.92 & 1.38 \\
	CR$_{k=6}$ & 6.68  & 2.52 & 3.94 & 1.50 \\
	CR$_{k=7}$ & 6.58 & 2.60 & 3.90 & 1.50 \\
	CR$_{k=8}$ & 6.56 & 2.48 & 3.64 & 1.54\\
	CR$_{k=9}$ & 6.59 & 2.69  & 3.84 & 1.60\\
	CR$_{k=10}$ & 6.32 & 2.52 & 3.61 & 1.62 \\
	\bottomrule
    \end{tabular}}
    \caption{\small{WER and CER values for \textbf{CR} with varying number of ConvNet layers trained using \emph{standard loss} function.}}
    \label{tab:ablation_layers}
\end{table}

\textbf{Loss Function.} The loss function in \S~\ref{subsec:loss_section} rewards higher the model's correcting behavior. Table~\ref{tab:ablation_loss} shows the ablation results for CR with varying $\lambda$ values for $\mathcal{L}'$ and fixed ConvNet layers ($k=1$ and $k=3$) as the best performing configurations in Table~\ref{tab:ablation_layers}. Here too due to the different characteristics of the evaluation scenarios, different $\lambda$ values are optimal for CR. We note that for \emph{eval--1}, a higher $\lambda$ of $0.3$ yields the best performance. This shows that for diverging train and test sets, e.g. \emph{eval--1}, the models need more stringent guidance in distinguishing copying from correcting behavior.

\begin{table}[th]
    \centering\small
    \begin{tabular}{l l l l l }
        \toprule
        & \multicolumn{2}{c}{\emph{eval--1}} & \multicolumn{2}{c}{\emph{eval--2}}\\
        \midrule
 & \textbf{WER} & \textbf{CER} & \textbf{WER} & \textbf{CER}\\
	\midrule
	CR$_{\lambda=0.1}$ & 6.22 & 2.16  & \textbf{3.59} & \textbf{1.31}\\
	CR$_{\lambda=0.2}$ & 6.31 &  2.17 & 3.79 & 1.42\\
	CR$_{\lambda=0.3}$ & \textbf{5.98} &  \textbf{2.07} & 4.24 & 1.51\\
	CR$_{\lambda=0.4}$ & 6.37 &  2.17 & 3.90 & 1.33\\
	CR$_{\lambda=0.5}$ & 6.37 &  2.16 & 3.83 & 1.45\\
	CR$_{\lambda=0.6}$ & 6.63 &  2.22 & 3.90 & 1.41\\ 
    \bottomrule
    \end{tabular}
    \caption{\small{WER and CER results for CR with different $\lambda$ for \emph{custom loss function}.}}
    \label{tab:ablation_loss}
\end{table}

\subsection{Page Level Performance}

Evaluation results in \S~\ref{subsec:model_performance} convey the ability of models to correct erroneous input at snippet level. However, there are challenges on applying post-hoc correction models on real-world OCR transcriptions, which do not have their textual content separated into coherent and non-overlapping snippets.

In this section, for our model \textbf{CR}, at \emph{page level} we assess the accuracy of undertaken actions in correcting the erroneous input text to its target form. Table~\ref{tab:page_level_action} shows the set of actions that a model can undertake. We carry out a manual evaluation on an out-of-corpus book (book code \texttt{Z168355305}), that is not present in our ground-truth, for which we randomly sample a set of 4 pages.

\begin{table}[h!]
    \centering\small
    \resizebox{0.9\columnwidth}{!}{%
    \begin{tabular}{l p{6cm}}
    \toprule
    \emph{action} &  \emph{description}\\
    \midrule
    \textbf{S} & accuracy of token segmentation\\
    \midrule
    \textbf{M} & accuracy of token merging\\
    \midrule
    \textbf{R} & accuracy of token character replacement (insertion/update/delete)\\
    \bottomrule
    \end{tabular}}
    \caption{\small{Page level actions are used to measure model's performance at page level.}}
    \label{tab:page_level_action}
\end{table}

We apply CR, namely its assess the accuracy of actions of correction during the decoding phase, over the OCR transcribed pages line by line with a window of 5 tokens. For each decoding step that produces an output that is \emph{different} from the input, we assess the accuracy of that action. Table~\ref{tab:manualpage} shows the precision of CR for the different set of actions for the different pages. The results show that CR is robust and can be applied without much change even at page level with high accuracy of post-hoc correction behavior.

\begin{table}[h!] \small
    \centering
    \resizebox{0.9\columnwidth}{!}{%
    \begin{tabular}{l l l l l}
        \toprule
      
  \emph{page}  & $\mathbf{S}$ & $\mathbf{M}$ & $\mathbf{R}$ & \emph{actions}\\
  \midrule 
  9 & 0.878 (66) & - & 0.737 (19) & 85\\
  10 & 0.976 (83) & - & 0.586 (29) & 112\\
  16 & 0.960 (50) & 0.0 (1) & 0.652 (23) & 73\\
  17 & 0.933 (90) & - & 0.621 (29) & 119\\

 \bottomrule
    \end{tabular}}
    \caption{\small{Precision for \textbf{S}, \textbf{M}, \textbf{R} actions. In brackets are the number of undertaken actions, and the rightmost column has all actions.}}
    \label{tab:manualpage}
\end{table}

% \begin{table}[h!] \small
% \resizebox{\columnwidth}{!}{%
%     \centering
%     \begin{tabular}{l l l l l l l }
%         \toprule
      
%   P_{id}  & $\mathbf{s}$  & $\mathbf{\hat{s}}$ & $\mathbf{m}$ & $\mathbf{\hat{m}}$ & $\mathbf{r}$ & $\mathbf{\hat{r}}$ \\
%   9 & 58 $(68.2\%)$ & 8 $(9.4\%)$ & - & - & 14 $(16.5\%)$ & 5 $(5.9\%)$ \\
%   10 & 81 (72.3\%) & 2 (1.8\%) & - & - & 17 (15.2\%) & 12 (10.7\%) \\
%   16 & 48 (64.9\%) & 2 (2.7\%) & - & 1 (1.3\%) & 15 (20.3\%) & 8 (10.8\%) \\
%   17 & 84 (70.6\%) & 6 (5\%) & - & - & 18 (15.1\%) & 11 (9.3\%) \\

%  \bottomrule
%     \end{tabular}}
%     \caption{Manual page evaluation}
%     \label{tab:manualpage}
% \end{table}

\subsection{Robustness}\label{subsec:robustness}
We conduct a robustness test of CR approach to check: (i) \emph{in-group} post-hoc correction performance, where test instances come from the same books as the training ones, and (ii) \emph{out-of-group}, where we train on one group and test on the rest of the groups. Table~\ref{tab:book_groups} shows the groups of books we use for (i) and (ii).
\begin{table}[ht!]
    \centering\small
    \resizebox{0.8\columnwidth}{!}{%
    \begin{tabular}{l l l l l}
    \toprule
         &  \#Train & \#Dev & \#Test & Book IDs \\ 
         \midrule
       \textbf{G1}  & 312k & 34.7k & 86.1k & (8, 5, 12, 11) \\
       \midrule
       \textbf{G2} & 58.9k & 6.5k & 17.2k & (2, 1, 3, 10) \\
       \midrule
       \textbf{G3} & 217.3k & 24k & 59.8k & (4, 7, 6, 9) \\
       \bottomrule
    \end{tabular}}
    \caption{\small{Book splits for assessing CR robustness.}}
    \label{tab:book_groups}
\end{table}

Table~\ref{tab:cross_val_lambda} shows the in-group and out-of-group post-hoc correction scores for CR when using a single ConvNet layer, using the standard and the custom loss functions, respectively. It can be seen that when the models are trained on a similar corpus (in-group), the error reduction is significantly higher compared to the evaluation on the out-of-group corpus. Furthermore, we note that the custom loss function, consistently provides better trained models for post-hoc correction.  
\begin{table}[ht!]
    \centering\small
    \resizebox{0.9\columnwidth}{!}{%
    \begin{tabular}{l l l l l l l}
        \toprule
        & \multicolumn{2}{c}{\textbf{\emph{G1}}} & \multicolumn{2}{c}{\textbf{\emph{G2}}}& \multicolumn{2}{c}{\textbf{\emph{G3}}}\\
        \midrule
   & \textbf{WER} & \textbf{CER} & \textbf{WER} & \textbf{CER} & \textbf{WER} & \textbf{CER}\\
	\midrule
	OCR & 28.1 & 5.7 & 34.0 & 7.1 & 31.2 & 5.8 \\
	\midrule
	\multicolumn{7}{c}{\emph{standard loss function}} \\
	\midrule
	\textbf{G1} & 10.1 & 2.9  & 24.7 & 6.3  & 18.9 & 4.9   \\
    \textbf{G2}  & 21.5 & 5.6  & 15.9 & 4.4  & 20.2 & 4.9 \\
    \textbf{G3}  & 16.9 & 4.4 & 18.9 & 4.8  & 10.4 & 2.6 \\
    \midrule
    \multicolumn{7}{c}{\emph{custom loss function}} \\
	\textbf{G1} & 10.1 & 2.8 & 24.2 & 5.6  & 18.4 & 4.4  \\
	\textbf{G2}  & 21.5 & 5.1 & 17.0 & 4.1  & 20.3  & 4.4 \\
	\textbf{G3}  & 16.9 & 4.3   & 18.7 & 4.6   & 10.3  & 2.5\\
    \bottomrule
    \end{tabular}}
    \caption{\small{CR results with $k=1$ trained using the standard and custom loss function with $\lambda = 0.1$}}
    \label{tab:cross_val_lambda}
\end{table}

The results in Table~\ref{tab:cross_val_lambda} show that CR is robust providing highly significant decrease in terms of WER and CER, with an average of WER decrease of 52\% for in-group with both the standard and custom loss. Whereas the out-of-group WER reduction is with 34\% and 35\% using the standard and custom loss, respectively. In terms of CER, for in-group we get a CER decrease of 47.6\% and 50\% for standard and custom loss, respectively. The advantage of the custom loss is shown for out-of-group evaluation, where the CER decrease is much more significant with 16.71\% for standard loss function compared to 23.3\% using the custom loss function.

From the three groups, when training on G3 the out-of-group post-hoc correction performance is the  highest. This shows that on historical corpora, depending on the initial OCR error rate and possibly the error types due to the book's characteristics impact significantly the correction performance.

\subsection{Error Analysis}\label{subsec:eval_loss_CR}
Here we analyze the structure of some typical errors that we fail to correct.

\paragraph{\textbf{Word Segmentation.}} In terms of over-segmentation, the importance of the ConvNet layers in CR is shown when compared against CH and CH$_{\lambda}$. Common word segmentation \emph{errors} for CH and CH$_{\lambda}$ are for example, \emph{``Jndem``} to \emph{``Jn``} and \emph{``dem``}, \emph{``Jedoch``} to \emph{``Je``} and \emph{``doch``}. \emph{``vorbey \hspace{0.5mm}\longs treichen``} to \emph{``vor bey\longs treichen``}. Most of these errors can be traced back to frequent constituents of the compound that exist in isolation too.

\textbf{Character error.} There are easy character errors such as \emph{``mein``} which is OCRed to \emph{``mcin``} and is fixed by all approaches. However, for some words like \emph{``lö\hspace{0.5mm}\longs cken }, models like CH and Transformer correct them to the right word \emph{``lö\hspace{0.5mm}\longs eten}. CR fails to do so due to some frequent character bigrams such as \emph{``ck``} that are very frequent in the dataset.

\subsection{Dataset Limitations}
The OCR quality can vary greatly across books, and from page to page. Based on manual inspection, we note that in some cases the WER can go well beyond 80\%. It is expected that in such cases that the post-hoc OCR correction will vary too. Other possible issues include competing spellings for the same word, which may cause the models to encode conflicting information, yet, for transcribing historical texts, language normalization (i.e. opting for one spelling) is not recommended, as the meaning of the texts may change.

\paragraph{\textbf{Language Evolution.}} There is a significant difference between \emph{eval--1} and \emph{eval--2} in terms of correction results. One explanation is due to the word spelling variations across centuries. Some examples include the substitution of single characters in words, which if not known would lead to systematic correction mistakes, e.g. $j \rightarrow i$, $v \rightarrow u$, \longs \hspace{0.2mm} $\rightarrow s$, \"a $\rightarrow \stackrel{e}{\,a}$. Accordingly, due to the missing information about the spelling change in \emph{eval--1}, the corresponding WER and CER rates are higher.

\section{Conclusion}
In this work we assessed several approaches towards post-hoc correction. We find out that OCR transcription errors are contextual, and a large set are due word-segmentation, followed by word-errors. Models like Transformers have limited utility in this task, as pre-training is difficult to undertake, given the scarcity of historical corpora. 

We proposed a OCR post-hoc correction approach for historical corpora, which provides flexible means to capturing various OCR transcription errors that are subject to \emph{language evolution}, \emph{typeface} and \emph{book layout} issues. Through our approach CR we achieve great WER reduction rates with 82\% and 89\% for \emph{eval--1} and \emph{eval--2} scenarios, respectively. 

Furthermore, ablation studies show that all the introduced components in CR yield consistent improvement over the competitors. Apart from post-hoc correction performance at snippet level, CR proved to be robust at page-level too, where the undertaken correction steps are highly accurate.

Finally, we construct a release a new dataset for post-hoc correction of historical corpora in German language, consisting of more than 850k parallel textual snippets, which can help facilitate research for historical and low-resource corpora.

% For future work, we will consider directions such as incorporation of syntax information and the possibility of transfer learning from modern German corpora for the task of OCR correction.

\section*{Acknowledgments.} This work was partially funded by Travelogues (DFG: 398697847 and  FWF: I 3795-G28).
\balance


\begin{thebibliography}{38}
\expandafter\ifx\csname natexlab\endcsname\relax\def\natexlab#1{#1}\fi

\bibitem[{Afli et~al.(2016)Afli, Qiu, Way, and
  Sheridan}]{DBLP:conf/lrec/AfliQWS16}
Haithem Afli, Zhengwei Qiu, Andy Way, and P{\'{a}}raic Sheridan. 2016.
\newblock \href
  {http://www.lrec-conf.org/proceedings/lrec2016/summaries/280.html} {Using
  {SMT} for {OCR} error correction of historical texts}.
\newblock In \emph{Proceedings of the Tenth International Conference on
  Language Resources and Evaluation {LREC} 2016, Portoro{\v{z}}, Slovenia, May
  23-28, 2016}. European Language Resources Association {(ELRA)}.

\bibitem[{Bahdanau et~al.(2015)Bahdanau, Cho, and Bengio}]{bahdanau2014neural}
Dzmitry Bahdanau, Kyunghyun Cho, and Yoshua Bengio. 2015.
\newblock \href {http://arxiv.org/abs/1409.0473} {Neural machine translation by
  jointly learning to align and translate}.
\newblock In \emph{3rd International Conference on Learning Representations,
  {ICLR} 2015, San Diego, CA, USA, May 7-9, 2015, Conference Track
  Proceedings}.

\bibitem[{Ballesteros et~al.(2015)Ballesteros, Dyer, and
  Smith}]{ballesteros2015improved}
Miguel Ballesteros, Chris Dyer, and Noah~A. Smith. 2015.
\newblock \href {https://doi.org/10.18653/v1/d15-1041} {Improved
  transition-based parsing by modeling characters instead of words with lstms}.
\newblock In \emph{Proceedings of the 2015 Conference on Empirical Methods in
  Natural Language Processing, {EMNLP} 2015, Lisbon, Portugal, September 17-21,
  2015}, pages 349--359. The Association for Computational Linguistics.

\bibitem[{Barbaresi(2016)}]{barbaresi2016bootstrapped}
Adrien Barbaresi. 2016.
\newblock \href
  {https://www.linguistics.rub.de/konvens16/pub/3\_konvensproc.pdf}
  {Bootstrapped {OCR} error detection for a less-resourced language variant}.
\newblock In \emph{Proceedings of the 13th Conference on Natural Language
  Processing, {KONVENS} 2016, Bochum, Germany, September 19-21, 2016},
  volume~16 of \emph{Bochumer Linguistische Arbeitsberichte}.

\bibitem[{Brill and Moore(2000)}]{brill2000improved}
Eric Brill and Robert~C. Moore. 2000.
\newblock \href {https://doi.org/10.3115/1075218.1075255} {An improved error
  model for noisy channel spelling correction}.
\newblock In \emph{38th Annual Meeting of the Association for Computational
  Linguistics, Hong Kong, China, October 1-8, 2000}, pages 286--293. {ACL}.

\bibitem[{Cho et~al.(2014)Cho, van Merrienboer, G{\"{u}}l{\c{c}}ehre, Bahdanau,
  Bougares, Schwenk, and Bengio}]{cho2014learning}
Kyunghyun Cho, Bart van Merrienboer, {\c{C}}aglar G{\"{u}}l{\c{c}}ehre, Dzmitry
  Bahdanau, Fethi Bougares, Holger Schwenk, and Yoshua Bengio. 2014.
\newblock \href {https://doi.org/10.3115/v1/d14-1179} {Learning phrase
  representations using {RNN} encoder-decoder for statistical machine
  translation}.
\newblock In \emph{Proceedings of the 2014 Conference on Empirical Methods in
  Natural Language Processing, {EMNLP} 2014, October 25-29, 2014, Doha, Qatar,
  {A} meeting of SIGDAT, a Special Interest Group of the {ACL}}, pages
  1724--1734. {ACL}.

\bibitem[{Chung et~al.(2016)Chung, Cho, and Bengio}]{chung2016character}
Junyoung Chung, Kyunghyun Cho, and Yoshua Bengio. 2016.
\newblock \href {https://doi.org/10.18653/v1/p16-1160} {A character-level
  decoder without explicit segmentation for neural machine translation}.
\newblock In \emph{Proceedings of the 54th Annual Meeting of the Association
  for Computational Linguistics, {ACL} 2016, August 7-12, 2016, Berlin,
  Germany, Volume 1: Long Papers}. The Association for Computer Linguistics.

\bibitem[{Cohn et~al.(2016)Cohn, Hoang, Vymolova, Yao, Dyer, and
  Haffari}]{DBLP:conf/naacl/CohnHVYDH16}
Trevor Cohn, Cong Duy~Vu Hoang, Ekaterina Vymolova, Kaisheng Yao, Chris Dyer,
  and Gholamreza Haffari. 2016.
\newblock \href {http://aclweb.org/anthology/N/N16/N16-1102.pdf} {Incorporating
  structural alignment biases into an attentional neural translation model}.
\newblock In \emph{{NAACL} {HLT} 2016, The 2016 Conference of the North
  American Chapter of the Association for Computational Linguistics: Human
  Language Technologies, San Diego California, USA, June 12-17, 2016}, pages
  876--885.

\bibitem[{Dauphin et~al.(2017)Dauphin, Fan, Auli, and
  Grangier}]{dauphin2017language}
Yann~N. Dauphin, Angela Fan, Michael Auli, and David Grangier. 2017.
\newblock \href {http://proceedings.mlr.press/v70/dauphin17a.html} {Language
  modeling with gated convolutional networks}.
\newblock In \emph{Proceedings of the 34th International Conference on Machine
  Learning, {ICML} 2017, Sydney, NSW, Australia, 6-11 August 2017}, volume~70
  of \emph{Proceedings of Machine Learning Research}, pages 933--941. {PMLR}.

\bibitem[{Dong and Smith(2018)}]{dong2018multi}
Rui Dong and David Smith. 2018.
\newblock \href {https://doi.org/10.18653/v1/P18-1220} {Multi-input attention
  for unsupervised {OCR} correction}.
\newblock In \emph{Proceedings of the 56th Annual Meeting of the Association
  for Computational Linguistics, {ACL} 2018, Melbourne, Australia, July 15-20,
  2018, Volume 1: Long Papers}, pages 2363--2372. Association for Computational
  Linguistics.

\bibitem[{Dreyer et~al.(2008)Dreyer, Smith, and Eisner}]{dreyer2008latent}
Markus Dreyer, Jason Smith, and Jason Eisner. 2008.
\newblock \href {https://www.aclweb.org/anthology/D08-1113/} {Latent-variable
  modeling of string transductions with finite-state methods}.
\newblock In \emph{2008 Conference on Empirical Methods in Natural Language
  Processing, {EMNLP} 2008, Proceedings of the Conference, 25-27 October 2008,
  Honolulu, Hawaii, USA, {A} meeting of SIGDAT, a Special Interest Group of the
  {ACL}}, pages 1080--1089. {ACL}.

\bibitem[{Farra et~al.(2014)Farra, Tomeh, Rozovskaya, and
  Habash}]{farra2014generalized}
Noura Farra, Nadi Tomeh, Alla Rozovskaya, and Nizar Habash. 2014.
\newblock \href {https://doi.org/10.3115/v1/p14-2027} {Generalized
  character-level spelling error correction}.
\newblock In \emph{Proceedings of the 52nd Annual Meeting of the Association
  for Computational Linguistics, {ACL} 2014, June 22-27, 2014, Baltimore, MD,
  USA, Volume 2: Short Papers}, pages 161--167. The Association for Computer
  Linguistics.

\bibitem[{Gehring et~al.(2017{\natexlab{a}})Gehring, Auli, Grangier, and
  Dauphin}]{gehring2016convolutional}
Jonas Gehring, Michael Auli, David Grangier, and Yann~N. Dauphin.
  2017{\natexlab{a}}.
\newblock \href {https://doi.org/10.18653/v1/P17-1012} {A convolutional encoder
  model for neural machine translation}.
\newblock In \emph{Proceedings of the 55th Annual Meeting of the Association
  for Computational Linguistics, {ACL} 2017, Vancouver, Canada, July 30 -
  August 4, Volume 1: Long Papers}, pages 123--135. Association for
  Computational Linguistics.

\bibitem[{Gehring et~al.(2017{\natexlab{b}})Gehring, Auli, Grangier, Yarats,
  and Dauphin}]{gehring2017convolutional}
Jonas Gehring, Michael Auli, David Grangier, Denis Yarats, and Yann~N. Dauphin.
  2017{\natexlab{b}}.
\newblock \href {http://proceedings.mlr.press/v70/gehring17a.html}
  {Convolutional sequence to sequence learning}.
\newblock In \emph{Proceedings of the 34th International Conference on Machine
  Learning, {ICML} 2017, Sydney, NSW, Australia, 6-11 August 2017}, volume~70
  of \emph{Proceedings of Machine Learning Research}, pages 1243--1252. {PMLR}.

\bibitem[{Hochreiter and Schmidhuber(1997)}]{hochreiter1997long}
Sepp Hochreiter and J{\"{u}}rgen Schmidhuber. 1997.
\newblock \href {https://doi.org/10.1162/neco.1997.9.8.1735} {Long short-term
  memory}.
\newblock \emph{Neural Computation}, 9(8):1735--1780.

\bibitem[{Kalchbrenner and Blunsom(2013)}]{kalchbrenner2013recurrent}
Nal Kalchbrenner and Phil Blunsom. 2013.
\newblock \href {https://www.aclweb.org/anthology/D13-1176/} {Recurrent
  continuous translation models}.
\newblock In \emph{Proceedings of the 2013 Conference on Empirical Methods in
  Natural Language Processing, {EMNLP} 2013, 18-21 October 2013, Grand Hyatt
  Seattle, Seattle, Washington, USA, {A} meeting of SIGDAT, a Special Interest
  Group of the {ACL}}, pages 1700--1709. {ACL}.

\bibitem[{Kim et~al.(2016)Kim, Jernite, Sontag, and Rush}]{kim2016character}
Yoon Kim, Yacine Jernite, David~A. Sontag, and Alexander~M. Rush. 2016.
\newblock \href
  {http://www.aaai.org/ocs/index.php/AAAI/AAAI16/paper/view/12489}
  {Character-aware neural language models}.
\newblock In \emph{Proceedings of the Thirtieth {AAAI} Conference on Artificial
  Intelligence, February 12-17, 2016, Phoenix, Arizona, {USA}}, pages
  2741--2749. {AAAI} Press.

\bibitem[{LeCun et~al.(1995)LeCun, Bengio et~al.}]{lecun1995convolutional}
Yann LeCun, Yoshua Bengio, et~al. 1995.
\newblock Convolutional networks for images, speech, and time series.
\newblock \emph{The handbook of brain theory and neural networks},
  3361(10):1995.

\bibitem[{Ling et~al.(2015)Ling, Trancoso, Dyer, and Black}]{ling2015character}
Wang Ling, Isabel Trancoso, Chris Dyer, and Alan~W. Black. 2015.
\newblock \href {http://arxiv.org/abs/1511.04586} {Character-based neural
  machine translation}.
\newblock \emph{CoRR}, abs/1511.04586.

\bibitem[{LPS()}]{lps_package}
LPS.
\newblock \url{https://biopython.org/DIST/docs/api/Bio.pairwise2-module.html}.

\bibitem[{Lund et~al.(2013)Lund, Kennard, and Ringger}]{lund2013combining}
William~B. Lund, Douglas~J. Kennard, and Eric~K. Ringger. 2013.
\newblock \href {https://doi.org/10.1117/12.2006228} {Combining multiple
  thresholding binarization values to improve {OCR} output}.
\newblock In \emph{Document Recognition and Retrieval XX, part of the
  IS{\&}T-SPIE Electronic Imaging Symposium, Burlingame, California, USA,
  February 5-7, 2013, Proceedings}, volume 8658 of \emph{{SPIE} Proceedings},
  page 86580R. {SPIE}.

\bibitem[{Lund et~al.(2014)Lund, Ringger, and Walker}]{lund2014well}
William~B. Lund, Eric~K. Ringger, and Daniel~David Walker. 2014.
\newblock \href {https://doi.org/10.1117/12.2042502} {How well does multiple
  {OCR} error correction generalize?}
\newblock In \emph{Document Recognition and Retrieval XXI, San Francisco,
  California, USA, February 5-6, 2014}, volume 9021 of \emph{{SPIE}
  Proceedings}, pages 90210A--90210A--13. {SPIE}.

\bibitem[{Lund et~al.(2011)Lund, Walker, and Ringger}]{lund2011progressive}
William~B. Lund, Daniel~David Walker, and Eric~K. Ringger. 2011.
\newblock \href {https://doi.org/10.1109/ICDAR.2011.303} {Progressive alignment
  and discriminative error correction for multiple {OCR} engines}.
\newblock In \emph{2011 International Conference on Document Analysis and
  Recognition, {ICDAR} 2011, Beijing, China, September 18-21, 2011}, pages
  764--768. {IEEE} Computer Society.

\bibitem[{\"ONB()}]{onb}
\"ONB.
\newblock \url{https://www.onb.ac.at/}.

\bibitem[{Rajaraman and Ullman(2011)}]{rajaraman2011mining}
Anand Rajaraman and Jeffrey~David Ullman. 2011.
\newblock \emph{Mining of massive datasets}.
\newblock Cambridge University Press.

\bibitem[{Reul et~al.(2018{\natexlab{a}})Reul, Springmann, Wick, and
  Puppe}]{DBLP:conf/das/ReulSWP18}
Christian Reul, Uwe Springmann, Christoph Wick, and Frank Puppe.
  2018{\natexlab{a}}.
\newblock \href {https://doi.org/10.1109/DAS.2018.30} {Improving {OCR} accuracy
  on early printed books by utilizing cross fold training and voting}.
\newblock In \emph{13th {IAPR} International Workshop on Document Analysis
  Systems, {DAS} 2018, Vienna, Austria, April 24-27, 2018}, pages 423--428.

\bibitem[{Reul et~al.(2018{\natexlab{b}})Reul, Springmann, Wick, and
  Puppe}]{DBLP:journals/corr/abs-1810-03436}
Christian Reul, Uwe Springmann, Christoph Wick, and Frank Puppe.
  2018{\natexlab{b}}.
\newblock \href {http://arxiv.org/abs/1810.03436} {State of the art optical
  character recognition of 19th century fraktur scripts using open source
  engines}.
\newblock \emph{CoRR}, abs/1810.03436.

\bibitem[{Sahin and Steedman(2018)}]{csahin2018character}
G{\"{o}}zde~G{\"{u}}l Sahin and Mark Steedman. 2018.
\newblock \href {https://doi.org/10.18653/v1/P18-1036} {Character-level models
  versus morphology in semantic role labeling}.
\newblock In \emph{Proceedings of the 56th Annual Meeting of the Association
  for Computational Linguistics, {ACL} 2018, Melbourne, Australia, July 15-20,
  2018, Volume 1: Long Papers}, pages 386--396. Association for Computational
  Linguistics.

\bibitem[{Schnober et~al.(2016)Schnober, Eger, Dinh, and
  Gurevych}]{schnober2016still}
Carsten Schnober, Steffen Eger, Erik{-}L{\^{a}}n~Do Dinh, and Iryna Gurevych.
  2016.
\newblock \href {https://www.aclweb.org/anthology/C16-1160/} {Still not there?
  comparing traditional sequence-to-sequence models to encoder-decoder neural
  networks on monotone string translation tasks}.
\newblock In \emph{{COLING} 2016, 26th International Conference on
  Computational Linguistics, Proceedings of the Conference: Technical Papers,
  December 11-16, 2016, Osaka, Japan}, pages 1703--1714. {ACL}.

\bibitem[{Schulz and Kuhn(2017)}]{schulz-kuhn-2017-multi}
Sarah Schulz and Jonas Kuhn. 2017.
\newblock \href {https://doi.org/10.18653/v1/d17-1288} {Multi-modular
  domain-tailored {OCR} post-correction}.
\newblock In \emph{Proceedings of the 2017 Conference on Empirical Methods in
  Natural Language Processing, {EMNLP} 2017, Copenhagen, Denmark, September
  9-11, 2017}, pages 2716--2726. Association for Computational Linguistics.

\bibitem[{Silfverberg et~al.(2016)Silfverberg, Kauppinen, and
  Lind{\'e}n}]{silfverberg2016data}
Miikka Silfverberg, Pekka Kauppinen, and Krister Lind{\'e}n. 2016.
\newblock \href {https://doi.org/10.18653/v1/W16-2406} {Data-driven spelling
  correction using weighted finite-state methods}.
\newblock In \emph{Proceedings of the {SIGFSM} Workshop on Statistical {NLP}
  and Weighted Automata}, pages 51--59, Berlin, Germany. Association for
  Computational Linguistics.

\bibitem[{Sutskever et~al.(2014)Sutskever, Vinyals, and
  Le}]{sutskever2014sequence}
Ilya Sutskever, Oriol Vinyals, and Quoc~V. Le. 2014.
\newblock \href
  {http://papers.nips.cc/paper/5346-sequence-to-sequence-learning-with-neural-networks}
  {Sequence to sequence learning with neural networks}.
\newblock In \emph{Advances in Neural Information Processing Systems 27: Annual
  Conference on Neural Information Processing Systems 2014, December 8-13 2014,
  Montreal, Quebec, Canada}, pages 3104--3112.

\bibitem[{Textarchiv()}]{DT}
Deutsches Textarchiv.
\newblock \url{http://www.deutschestextarchiv.de/}.

\bibitem[{TK()}]{transformer_keras}
TK.
\newblock
  \url{https://github.com/Lsdefine/attention-is-all-you-need-keras/blob/master/transformer.py}.

\bibitem[{Vaswani et~al.(2017)Vaswani, Shazeer, Parmar, Uszkoreit, Jones,
  Gomez, Kaiser, and Polosukhin}]{vaswani2017attention}
Ashish Vaswani, Noam Shazeer, Niki Parmar, Jakob Uszkoreit, Llion Jones,
  Aidan~N. Gomez, Lukasz Kaiser, and Illia Polosukhin. 2017.
\newblock \href {http://papers.nips.cc/paper/7181-attention-is-all-you-need}
  {Attention is all you need}.
\newblock In \emph{Advances in Neural Information Processing Systems 30: Annual
  Conference on Neural Information Processing Systems 2017, 4-9 December 2017,
  Long Beach, CA, {USA}}, pages 5998--6008.

\bibitem[{Wang et~al.(2014)Wang, Xu, Li, and Zhang}]{wang2014probabilistic}
Ziqi Wang, Gu~Xu, Hang Li, and Ming Zhang. 2014.
\newblock \href {https://doi.org/10.1109/TKDE.2013.11} {A probabilistic
  approach to string transformation}.
\newblock \emph{{IEEE} Trans. Knowl. Data Eng.}, 26(5):1063--1075.

\bibitem[{Xie et~al.(2016)Xie, Avati, Arivazhagan, Jurafsky, and
  Ng}]{xie2016neural}
Ziang Xie, Anand Avati, Naveen Arivazhagan, Dan Jurafsky, and Andrew~Y. Ng.
  2016.
\newblock \href {http://arxiv.org/abs/1603.09727} {Neural language correction
  with character-based attention}.
\newblock \emph{CoRR}, abs/1603.09727.

\bibitem[{Xu and Smith(2017)}]{xu2017retrieving}
Shaobin Xu and David~A. Smith. 2017.
\newblock \href {https://doi.org/10.1109/JCDL.2017.7991587} {Retrieving and
  combining repeated passages to improve {OCR}}.
\newblock In \emph{2017 {ACM/IEEE} Joint Conference on Digital Libraries,
  {JCDL} 2017, Toronto, ON, Canada, June 19-23, 2017}, pages 269--272. {IEEE}
  Computer Society.

\end{thebibliography}
\end{document}